\documentclass[10pt,twocolumn,letterpaper]{article}
\usepackage{iccv}
\usepackage{times}
\usepackage{epsfig}
\usepackage{graphicx}
\usepackage{amsmath}
\usepackage{amssymb}
\usepackage{subfigure}
\usepackage{adjustbox}
\usepackage{caption}
\usepackage{multirow}
\usepackage{enumitem}
\usepackage{array}
\newcolumntype{H}{>{\setbox0=\hbox\bgroup}c<{\egroup}@{}}
\newcommand{\tabincell}[2]{\begin{tabular}{@{}#1@{}}#2\end{tabular}}
\usepackage{amssymb}% http://ctan.org/pkg/amssymb
\usepackage{pifont}% http://ctan.org/pkg/pifont
\newcommand{\cmark}{\ding{51}}%
\newcommand{\xmark}{\ding{55}}%
\newcommand{\eqspace}{-3.5pt}
\newcommand{\subsecspace}{-1.2mm}
\usepackage[pagebackref=true,breaklinks=true,letterpaper=true,colorlinks,bookmarks=false]{hyperref}
\iccvfinalcopy % *** Uncomment this line for the final submission

\ificcvfinal\fi

\begin{document}
	
	%%%%%%%%% TITLE
	\title{Self-Conditioned Probabilistic Learning of Video Rescaling}
	
	\author{\normalsize   Yuan Tian$^{1}$
		~~
		Guo Lu$^{2}$
		~~
		Xiongkuo Min$^{1}$
		~~
		Zhaohui Che$^{1}$
		~~
		Guangtao Zhai$^{1}$\thanks{Corresponding author.}
%		\and \normalsize
		~~
		Guodong Guo$^{3}$
		~~
		Zhiyong Gao$^{1}$
		\\
		\normalsize $^{1}$Shanghai Jiao Tong Unversity~~~~$^{2}$Beijing Institute of Technology~~~~$^{3}$Baidu\\
		{\tt\small \{ee\_tianyuan, minxiongkuo, chezhaohui, zhaiguangtao, zhiyong.gao\}@sjtu.edu.cn}\\
		{\tt\small guo.lu@bit.edu.cn, guoguodong01@baidu.com}		
	}

	\maketitle
	\ificcvfinal\thispagestyle{empty}\fi
	
	%%%%%%%%% ABSTRACT
	\begin{abstract}
		Bicubic downscaling is a prevalent technique used to reduce the video storage burden or to accelerate the downstream processing speed.
		However, the inverse upscaling step is non-trivial, and the downscaled video may also deteriorate the performance of downstream tasks.
		In this paper, we propose a
		self-conditioned probabilistic framework for video rescaling to learn the paired downscaling and upscaling procedures simultaneously.
		During the training, we decrease the entropy of the information lost in the downscaling by maximizing its probability conditioned on the strong spatial-temporal prior information within the downscaled video.
		After optimization, the downscaled video by our framework preserves more meaningful information, which is beneficial for both the upscaling step and the downstream tasks, \textit{e.g.}, video action recognition task. We further extend the framework to a lossy video compression system, in which a gradient estimator for non-differential industrial lossy codecs is proposed for the end-to-end training of the whole system.
		Extensive experimental results demonstrate the superiority of our approach on video rescaling, video compression, and efficient action recognition tasks.
	\end{abstract}
	
	%%%%%%%%% BODY TEXT
	\vspace{-6mm}
	\section{Introduction}
	
	High-resolution videos are widely used over various computer vision tasks~\cite{simonyan2014two}\cite{xu2015discriminative}\cite{fan2016video}\cite{lu2019deep}\cite{lu2018deep}\cite{tian2019video}\cite{yan2021dehib}. 
	However, considering the increased storage burden or the high computational cost,
	it is usually required to first downscale the high-resolution videos.
	Then we can either compress the output low-resolution videos for saving storage cost or feed them to the downstream tasks to reduce the computational cost. 
	Despite that this paradigm is prevalent, it has the following two disadvantages. First, it is non-trivial to restore the original high-resolution videos from the (compressed) low-resolution videos,
	even we use the latest super-resolution methods~\cite{liu2013bayesian,xue2019video,tao2017detail,wang2019edvr}. Second, it is also a challenge for the downstream tasks to achieve high performance based on these low-resolution videos.
	Therefore, it raises the question that whether the downscaling operation can facilitate the reconstruction of the high-resolution videos and also preserve the most meaningful information.
	
	Recently, this question has been partially studied as a single image rescaling problem~\cite{kim2018task,li2018learning,sun2020learned,xiao2020invertible},
	which learns the down/up scaling operators jointly.
	However, how to adapt these methods from image to video domain and leverage the rich temporal information within videos are still open problems.
	More importantly, modeling the lost information during downscaling is non-trivial.
	Current methods either ignore the lost information~\cite{kim2018task,li2018learning,sun2020learned} or assume it as an independent distribution in the latent space~\cite{xiao2020invertible},
	while neglecting the \textit{internal relationship} between the downscaled image and the lost information.
	Besides, all literature above has not explored how to apply the rescaling technique to the lossy image/video compression.
	
	In this paper, we focus on building a video rescaling framework and propose a self-conditioned probabilistic learning approach to learn a pair of video downscaling and upscaling operators by exploiting the information dependency within the video itself.
	Specifically,
	we first design a learnable frequency analyzer to decompose the original high-resolution video into its downscaled version
	and the corresponding high-frequency component.
	Then, a Gaussian mixture distribution is leveraged to model the high-frequency component by \textit{conditioning} on the downscaled video.
	For accurate estimation of the distribution parameters,
	we further introduce the local and global temporal aggregation modules to fuse the spatial information from adjacent downscaled video frames.
	Finally, the original video can be restored by a frequency synthesizer from the downscaled video and the high-frequency component sampled from the distribution.
	We integrate the components above as a novel self-conditioned video rescaling framework termed \textbf{SelfC} and optimize it by minimizing the negative log-likelihood for the distribution.
	
	Furthermore, we apply our proposed SelfC in two practical applications, \textit{i.e.}, lossy video compression and video action recognition. In particular, to integrate our framework with the existing non-differential video codecs (\textit{e.g.}, H.264~\cite{wiegand2003overview} and H.265~\cite{sullivan2012overview}), we propose an efficient and effective one-pass optimization strategy based on the \textit{control variates} method and approximate the gradients of traditional codecs in the back-propagation procedure, which formulates an end-to-end optimization system.

	Experimental results demonstrate that the proposed framework achieves state-of-the-art performance on the video rescaling task.
	More importantly, we further demonstrate the effectiveness of the framework in practical applications. For the lossy video compression task, compared with directly compressing the high-resolution videos, the video compression system based on our SelfC framework cuts the storage cost significantly (up to 50\% reduction). For the video action recognition task, our framework reduces more than 60\% computational complexity with negligible performance degradation. 
	In summary, our main contributions are:
	\vspace{-2mm}
	\begin{itemize}
		\setlength{\itemsep}{0pt}
		\setlength{\parsep}{0pt}
		\setlength{\parskip}{0pt}
		\item We propose a probabilistic learning framework dubbed \textbf{SelfC} for the video rescaling task, which models the lost information during downscaling as a dynamic distribution conditioned on the downscaled video.
		\item Our approach exploits rich temporal information in downscaled videos for an accurate estimation of the distribution parameters
		by introducing the specified local and global temporal aggregation modules.
		\item We propose a gradient estimation method for non-differential lossy codecs based on the control variates method and Monte Carlo sampling technique, extending the framework to a video compression system.
	\end{itemize}
	
	\vspace{-4mm}
	\section{Related Work}
	
	\textbf{Video Upscaling after Downscaling.}
	Previous SR works~\cite{liu2013bayesian,xue2019video,tao2017detail,wang2019edvr,haris2019recurrent,jo2018deep} mainly
	leverage a heavy neural network to hallucinate the lost details during the downscaling, only achieving unsatisfactory results.
	Taking the video downscaling method into consideration may help mitigate the ill-posedness of the video upscaling procedure.
	
	There are already a few works on single image rescaling task following this spirit.
	For example, Kim \etal~\cite{kim2018task} proposed a task-aware downscaling model based on an auto-encoder framework.
	Li \etal~\cite{li2018learning} proposed to use a neural network to estimate the downscaled low-resolution images for a given super-resolution method.
	Yang \etal~\cite{yang2020superpixel} proposed a superpixel-based downsampling/upsampling scheme to effectively preserve object boundaries.
	Recently, Xiao \etal~\cite{xiao2020invertible} proposed to leverage a deep invertible neural network (INN) to model the problem, which maps the complex distribution of the lost information to an \textit{independent} and \textit{fixed} normal distribution.
	
	However, these methods neither leverage the temporal information between adjacent frames, which is important for video-related tasks,
	nor consider the fact that the components of
	different frequencies in natural images or videos are conditionally dependent~\cite{wainwright1999scale,strela2000image,portilla2003image,wainwright2001random}.
	
	\textbf{Video Compression.}
	Several traditional video compression algorithms have been proposed and widely deployed, such as H.264~\cite{wiegand2003overview} and H.265~\cite{sullivan2012overview}.
	Most of them follow the predictive coding architecture and rely on the sophisticated hand-crafted transformations to analyze the redundancy within the videos.
	Recently, fully end-to-end video codecs~\cite{lu2019dvc}~\cite{habibian2019video}\cite{djelouah2019neural}\cite{yang2020learning}\cite{agustsson2020scale}\cite{lin2020m}\cite{lu2020content}\cite{hu2021fvc}\cite{lu2020end} have been proposed by optimizing the rate-distortion trade-off. They demonstrate promising performance and may be further improved by feeding more ubiquitous videos in the wild. However, they haven't been widely used by industrial due to computational inefficiency.	
	In contrast, our framework can be readily integrated with the best traditional video codecs and further saves the video storage space significantly.

	\textbf{Video Action Recognition.}
	Simonyan \etal~\cite{simonyan2014two} first proposed the two-stream framework.
	Feichtenhofer \etal~\cite{feichtenhofer2016convolutional} then improved it.
	Later, wang \etal~\cite{wang2018temporal} proposed a new sparse frame sampling strategy.
	Recently, 3D networks~\cite{tran2015learning}\cite{carreira2017quo}\cite{hara2017learning}\cite{tran2018closer}\cite{qiu2017learning}\cite{feichtenhofer2019slowfast}\cite{tian2020self} also show promising performance.
	Our work can accelerate the off-the-shelf action CNNs by 3-4 times while reserving the comparable performance.
	We mainly conduct experiments on light-weight 2D action CNNs (\textit{e.g.}, TSM~\cite{lin2019tsm}) for efficiency.
    	
	\begin{figure*}[!th] 
	\vspace{-5mm}
		\centering
		\newcommand{\widthscalefive}{0.3}
		\includegraphics[width=0.96 \textwidth]{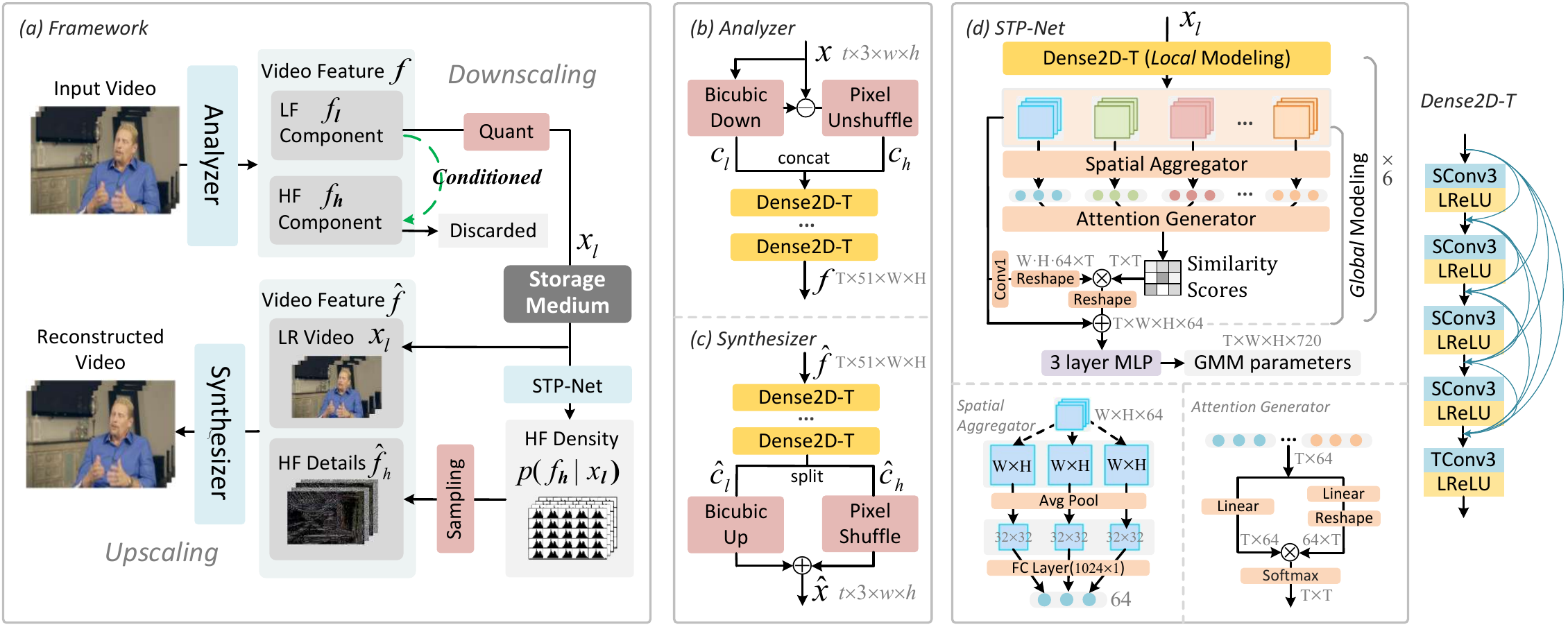}
		\caption{
		    Overview of the proposed SelfC framework. We exploit the conditional relationship between different frequency components within the input video for better learning of video rescaling.
%		    which are disentangled by the frequency analyzer.
		    During downscaling, the low-frequency (LF) component $f_l$ is quantized to produce the low-resolution (LR) video $x_l$ and then stored. The storage medium can be lossless or lossy for different applications.
			During upscaling, the probability density $p(f_h|x_l)$ of the high-frequency (HF) component $f_h$ is predicted by a spatial-temporal prior network (\textsf{\small STP-Net}) from $x_l$.
			Then, $x_l$ and the sampled HF details $\hat{f}_h$ are reconstructed to the high-resolution video by the frequency synthesizer.
			We also introduce some new symbols, W$ = w/k$, H$ = h/k$ and T$ = t$, for better indication of the tensor dimensions.
			\textsf{\small Conv1/SConv3/TConv3} denote the 3D convolutions of kernel size $1\times1\times1$/$1\times3\times3$/$3\times1\times1$.
			LReLU denotes the Leaky-ReLU non-linearity~\cite{xu2015empirical}.
		}
		\vspace{-3mm}
		\label{fig:framework}
	\end{figure*} 
	
	%\vspace{-3mm}
	\section{Proposed Method}
	%\vspace{-2mm}
    An overview of our proposed SelfC framework is shown in Fig.~\ref{fig:framework} (a).
	During the downscaling procedure, given a high-resolution (HR) video,
	a frequency analyzer (FA) (Section~\ref{sec:fa}) first converts it into video features $f$, where the first $3$ channels are low-frequency (LF) component $f_{l}$, the last $3 \cdot k^2$ channels are high-frequency (HF) component $f_{h}$, and
	%that is discarded.
	$k$ is the downscaling ratio.
	Then, $f_{l}$ is quantized to a LR video $x_l$ for storage.
	$f_h$ is discarded in this procedure.
	
	During the upscaling procedure, given the LR video $x_l$, the spatial-temporal prior network (STP-Net) (Section~\ref{sec:HFG}) predicts the probability density function of the HF component $f_h$:
%	\vspace{\eqspace}
	\begin{equation}
	\label{eq:prob_infer}
	p(f_h|x_l) = \textsf{\small STP-Net}(x_l).
%	\vspace{\eqspace}
	\end{equation}
	
	We model $p(f_h|x_l)$ as a continuous mixture of the parametric Gaussian distributions (Section~\ref{sec:prob_model}).
	Then, a case of the HF component $\hat{f}_h$ related to LR video $x_l$ is drawn from the distribution.
	Finally, we reconstruct the HR video from the concatenation of HF component $\hat{f}_h$ and LR video $x_l$ by the frequency synthesizer (FS).

	\subsection{Frequency Analyzer and Synthesizer}
	\vspace{\subsecspace}
	\label{sec:fa}
	As shown in Fig.~\ref{fig:framework} (b),
	we first decompose the HR input video $x$ as the $k$ times downscaled low-frequency component $c_{l} := \textsf{\small Down}(x) \in \mathbb{R}^{t\times 3 \times \frac{h}{k} \times \frac{w}{k}}$ and the residual high-frequency component $c_{h} := \textsf{\small PixelUnshuffle}(x - \textsf{\small Up}(c_{l})) \in \mathbb{R}^{t\times 3 \cdot k^2 \times \frac{h}{k} \times \frac{w}{k}}$,
	where $h\times w$ denotes the spatial scale of the original video and $t$ denotes the video length. $\textsf{\small Down}$ and $\textsf{\small Up}$ represent bicubic downscaling and upscaling operations with the scaling ratio $k$. $\textsf{\small PixelUnshuffle}$ is the inverse operation of the pixel shuffling operation proposed in ~\cite{shi2016real}, where the scaling ratio is also $k$.
	Then, we use a learnable operation $\mathcal{T}$ to transform $c_l$ and $c_h$ to the output features $f := \mathcal{T}(c_{l} \textsf{\small \textcopyright} c_{h})$,
	where $ \textsf{\small \textcopyright}$ denotes the channel concatenation operation. 
	The produced video feature $f$ consists of LF component $f_l$ and HF component $f_h$. 
	The network architecture for $\mathcal{T}$ is very flexible in our framework and we use multiple stacking Dense2D-T blocks to implement it by default.
	The Dense2D-T block is modified from the vanilla Dense2D block~\cite{iandola2014densenet} by replacing the last spatial convolution with the temporal convolution.

	The architecture of the frequency synthesizer is symmetric with the analyzer, as shown in Fig.~\ref{fig:framework} (c). Specifically,
	we use channel splitting, bicubic upscaling, and pixel shuffling operations to synthesize the final high-resolution videos based on the reconstructed video feature $\hat{f}$.

	\vspace{\subsecspace}
	\subsection{A Self-conditioned Probabilistic Model}
		\vspace{\subsecspace}
	\label{sec:prob_model}
	Directly optimizing $p(f_h|x_l)$ in Eq.~(\ref{eq:prob_infer}) through gradient descent is unstable due to the unsmooth gradient~\cite{kim2018task} of the quantization module.
	Thus, we optimize $p(f_h|f_l)$ instead during the training procedure.
	Specifically, we represent the high-frequency component $f_h$ as a continuous multi-modal probability distribution conditioned on the low-frequency component $f_l$, which is formulated as:
	\vspace{\eqspace}
	\begin{equation}
	\label{eq:prob_product}
	p(f_{h}|f_{l}) = \prod_{o}^{} p(f_{h}(o) | f_{l}),
	\vspace{\eqspace}
	\end{equation}
	where $o$ denotes the spatial-temporal location.
	We use a continuous Gaussian Mixture Model (GMM)~\cite{reynolds2009gaussian} to approximate $p$ with component number $K=5$.
	The distributions are defined by the learnable mixture weights $w_{o}^k$, means $\mu_{o}^k$ and log variances $\sigma_{o}^k$.
	With these parameters, the distributions can be accurately determined as:
	\vspace{\eqspace}
	\begin{equation}
	p(f_{h}(o) | f_{l})  = \sum_{k=1}^{K} w_{o}^k p_g (f_{h}(o)|\mu_{o}^k, e^{\sigma_{o}^k}),
	\vspace{\eqspace}
	\end{equation}
	where
	\vspace{\eqspace}
	\begin{equation}
	p_g(f|\mu, \sigma^2) = \frac{1}{\sqrt{\pi}\sigma}e^{-\frac{(f-\mu)^2}{\sigma^2}}.
	\vspace{\eqspace}
	\end{equation}
	\vspace{\subsecspace}
	\subsection{Spatial-temporal Prior Network (STP-Net)}
		\vspace{\subsecspace}
	\label{sec:HFG}
	As shown in Fig.~\ref{fig:framework} (d), to estimate the parameters of the distribution above, we propose the STP-Net to model both the local and global temporal information within the downscaled video.
	We first utilize the Dense2D-T block to extract the short-term spatial-temporal features for each input frame.
	In this stage, only information from local frames, \textit{i.e.}, the previous or the next frames, are aggregated into the current frame, while the temporally long-range dependencies in videos are neglected. Therefore, we further introduce the attention mechanism for modeling the global temporal information.
	More specifically, the spatial dimension of the short-term spatial-temporal features is first reduced by a spatial aggregator, which is implemented as an average pooling operation followed by a full-connected (FC) layer. The output scale of the pooling operation is 32$\times$32. Then we use the dot-producting operation to generate the attention map, which represents the similarity scores between every two frames. Finally, we refine the local spatial-temporal features based on the similarity scores.
	We repeat the following procedure six times to extract better video features.
    After that, a three-layer multi-layer perceptron (MLP) is used to estimate the parameters of the GMM distribution.
	\vspace{\subsecspace}
	\subsection{Quantization and Storage Medium}
		\vspace{\subsecspace}
	We use rounding operation as the quantization module, and store the output LR videos by lossless format, \textit{i.e.}, H.265 lossless mode.
	The gradient of the module is calculated by the Straight-Through Estimator~\cite{bengio2013estimating}.
	We also discuss how to adapt the framework to more practical lossy video formats such as H.264 and H.265 in Section~\ref{app1_compression}.
	\vspace{\subsecspace}
	\subsection{Training Strategy}
		\vspace{\subsecspace}
	\label{sec:fw-learn}
	 Building a learned video rescaling framework is non-trivial, especially the generated low-resolution videos are expected to benefit both the upscaling procedure and the downstream tasks. We consider the following objectives.

	\textbf{Learning the self-conditioned probability.}
	First, to make sure the STP-Net can obtain an accurate estimation for the HF component $f_h$,
	we directly minimize the negative log-likelihood of $p(f_{h}|f_{l})$ in Eq.~(\ref{eq:prob_product}):
	\vspace{\eqspace}
%	\vspace{-1mm}
	\begin{equation}
	\label{eq:conditon_model}
	\mathcal{L}_{c} = -\sum_{i=0}^{N}log(p(f_{h}^i | f_{l}^i)),
	\vspace{\eqspace}
%	\vspace{-1mm}
	\end{equation}
	where $N$ is the number of the training samples.
	
	\textbf{Mimicking Bicubic downscaling.}
Then the downscaled video is preferred to be similar to the original video, making its deployment for the downstream tasks easier. Therefore, we regularize the downscaled video before quantization, \textit{i.e.}, $f_l$, to mimic the bicubic downsampled $x$:
%	\vspace{\eqspace}
	\begin{equation}
	\mathcal{L}_{mimic} =  ||x_{bicubic} - f_{l}||_{2}, x_{bicubic} = \textsf{\small Bicubic}(x).
%	\vspace{\eqspace}
	\end{equation}

	\textbf{Penalizing $\mathcal{T}$.}
		Without any extra constraint, 
	Eq.~(\ref{eq:conditon_model}) can be easily minimized by tuning $f_{h}$ to one constant tensor for any input video.
	Thus, to avoid the trivial solution, 
% 	The information lost during the transformation
the CNN parts of the frequency analyzer and synthesizer are penalized by the photo-parametric loss (\textit{i.e.}, $\ell_2$ loss) between the video directly reconstructed from $f$ and the original input $x$:
%	\vspace{\eqspace}
%	\vspace{-1mm}
	\begin{equation}
	\label{eq:loss_pen}
	\mathcal{L}_{pen} = || x - \textsf{\small FS}(f) ||_{2}.
%	\vspace{\eqspace}
	\end{equation}

	\textbf{Minimizing reconstruction difference.}
	Finally, the expected difference between the reconstructed video sampled from the model and the original video should be minimized:
	\begin{equation}
	\label{eq:lf_recons}
	\mathcal{L}_{recons} = \ell (x,\hat{x}), \hat{x} = \textsf{\small FS}(x_l  \textsf{\small \textcopyright} \hat{f}_{h}),
	\end{equation}
	where
% 	$\operatorname{FS}$ denotes the frequency synthesizer, 
	$\ell$ denotes a photo-metric loss (\textit{i.e.}, $\ell_1$ loss), $ \textsf{\small \textcopyright}$ denotes the channel-wise concatenation operation.
	In each training iteration, $\hat{f}_{h}$ is sampled from the distribution constructed from the parameters output by STP-Net, conditioning on the LR video $x_l$.
	To enable an end-to-end optimization, we apply the ``reparametrization trick''~\cite{kingma2013auto,rezende2014stochastic,graves2016stochastic} to make the sampling procedure differentiable.

	The total loss is then given by:
%	\vspace{\eqspace}
	\begin{equation}
	\label{eq:loss_lowlevel}
% 	\resizebox{.88\hsize}{!}{
	\mathcal{L}_{selfc} = \lambda_1 \mathcal{L}_{c} + \lambda_2 \cdot k^2 \mathcal{L}_{mimic} + \lambda_3 \mathcal{L}_{pen} + \lambda_4 \mathcal{L}_{recons},
% 	}
%	\vspace{\eqspace}
	\end{equation}
	where $\lambda_1$, $\lambda_2$, $\lambda_3$ and $\lambda_4$ are the balancing parameters, and $k$ is the scaling ratio.
	The loss function of our framework may seem a little bit complicated. However, we want to mention that the performance of our framework is not sensitive to these hyper-parameters, and directly setting all the parameters to 1 already achieves reasonable performance.
	
	\vspace{\subsecspace}
	\subsection{Application I: Video Compression}
\begin{figure}[!b] 
				\vspace{-3mm}
		\centering
		\newcommand{\widthscalefive}{0.3}
		\includegraphics[width=0.48 \textwidth]{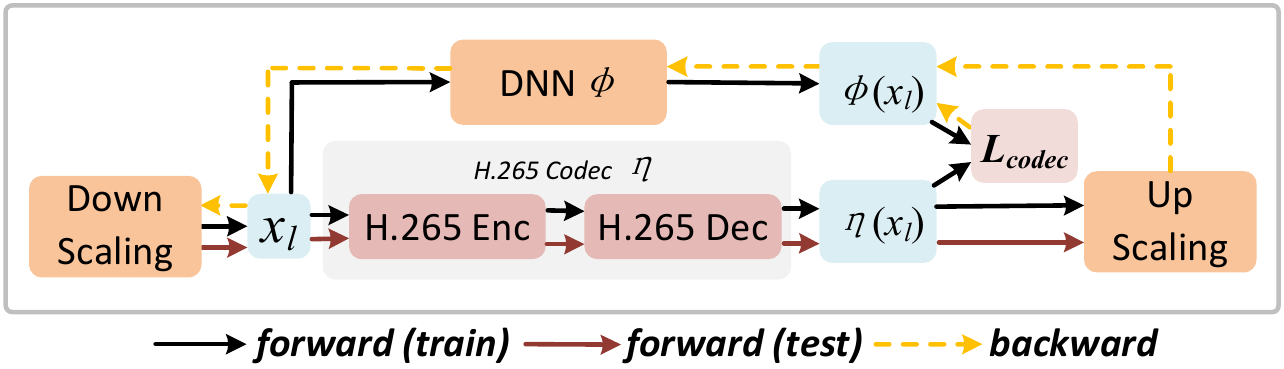}
		\vspace{-6mm}
		\caption{
			We introduce a surrogate DNN to calculate the gradient of the non-differential codec. We take the H.265 codec as an example.
			%			 though our method can be applied to arbitrary traditional codecs. Please refer to Section~\ref{app1_compression} for more details.
		}
		\vspace{-5mm}
		\label{fig:codec_framework}
	\end{figure}

	\vspace{\subsecspace}
	\label{app1_compression}
	In this section, we extend the proposed SelfC framework to a lossy video compression system, which aims to demonstrate the effectiveness of our approach in reducing the video storage space.
	The whole system is shown in Fig.~\ref{fig:codec_framework}. Specifically, we first use the SelfC to generate the downscaled video $x_l$, which will be compressed by using the off-the-shelf industrial video codecs, \textit{e.g.}, H.265 codec. Then at the decoder side, the compressed videos will be decompressed and upscaled to the full resolution video. 

	Considering the traditional video codecs are non-differential, we further propose a novel optimization strategy. 
	Specifically, we introduce a differentiable surrogate video perturbator $\phi$, which is implemented as a deep neural network (DNN) consisting of 6 Dense2D-T blocks. During the back-propagation stage, the gradient of the codec can be approximated by that of $\phi$, which is tractable. During the test stage, the surrogate DNN is removed and we directly use the H.265 codec for compression and decompression.

	According to the \textit{control variates} theory~\cite{glynn2002some,grathwohl2017backpropagation}, $\phi$ can be a low-variance gradient estimator for the video codec (\textit{i.e.,} $\eta$) when (1) the differences between the outputs of the two functions are minimized and (2) the correlation coefficients $\rho$ of the two output distributions are maximized. 
	
	Therefore, we introduce these two constraints to the optimization procedure of the proposed SelfC based video compression system. The loss function for the surrogate video perturbator is formulated as:
% 	The two condition can be mathematically formulated as the following loss function:
%	\vspace{\eqspace}
	\begin{equation}
	\mathcal{L}_{codec} = || \eta(x_l) - \phi(x_l) ||_2 - \lambda_\rho \rho(\eta,\phi),
%	\vspace{\eqspace}
	\end{equation}
	where $\lambda_\rho$ is set to a small value, \textit{i.e.}, $10^{-5}$,
	and $\rho$ is estimated within each batch by Monte Carlo sampling:
%	\vspace{\eqspace}
	\begin{equation}\small
% 	\resizebox{.87\hsize}{!}{
	\rho(\eta,\phi) = \frac{\sum_{k=1}^{N}(\eta(x_l^k) - \mathbb{E}[\eta] ) (\phi(x_l^k) - \mathbb{E}[\phi] ) }
	{\sqrt{\sum_{k=1}^{N}(\eta(x_l^k) - \mathbb{E}[\eta])^2} \sqrt{\sum_{k=1}^{N}(\phi(x_l^k) - \mathbb{E}[\phi])^2} },
% 	}
%	\vspace{\eqspace}
	\end{equation}
	where 
%	\vspace{\eqspace}
%	\vspace{-4mm}
	\begin{equation}
	\mathbb{E}[\eta] = \frac{1}{N} \sum_{k=1}^{N}\eta(x_l^k),~ \mathbb{E}[\phi] = \frac{1}{N} \sum_{k=1}^{N}\phi(x_l^k),
%	\vspace{\eqspace}
	\end{equation}
	and $N$ denotes the batch size.
	Note that $\rho$ is separately computed on each spatial-temporal location and then averaged.
	
	Finally, the total loss function for the SelfC based video compression system is given by:
	\vspace{\eqspace}
	\begin{equation}
	\mathcal{L}_{compression} = \mathcal{L}_{selfc} + \lambda_{codec} \mathcal{L}_{codec},
%	\vspace{\eqspace}
	\end{equation}
	where $\lambda_{codec}$ is the balancing weight.

	\vspace{\subsecspace}
	\subsection{Application II: Efficient Action Recognition}
		\vspace{\subsecspace}
	We further apply the proposed SelfC framework to the video action recognition task. 
	Specifically, we adopt the LR videos (\textit{i.e.}, $x_l$) downscaled by our framework as the input of action recognition CNNs for efficient action recognition.
	Considering the downscaler of our approach can preserve meaningful information for the downstream tasks and the complexity of itself can be rather low, inserting the downscaler before the off-the-shelf action CNNs can reduce the huge computational complexity of them with negligible performance drop. 
	Moreover, the light-weightiness of the rescaling framework makes the joint optimization tractable. In fact, compared with bicubic downscaling operation, our downscaler in SelfC framework can still generate more informative low-resolution videos for the action recognition task even without the joint training procedure. Please see Section~\ref{sec_exp_action} for more experimental results.
	
	\vspace{-2mm}

	\section{Experiments}
	\vspace{\subsecspace}
	\subsection{Dataset}
	\vspace{\subsecspace}
	We use the videos in Vimeo90K dataset~\cite{xue2019video} as our training data, which are also adopted by the recent video super resolution methods~\cite{RSDN,TGA,wang2019edvr} and learnable video compression methods~\cite{lu2019dvc,hu2020improving,lu2020content}.
	\textit{For video rescaling task}, the evaluation datasets are the test set of Vimeo90K (denoted by Vimeo90K-T), the widely-used Vid4 benchmark~\cite{liu2013bayesian} and SPMCs-30 dataset~\cite{tao2017detail}.
	\textit{For video compression task}, the evaluation datasets include UVG~\cite{UVG}, MCL-JCV~\cite{wang2016mcl}, and HEVC Class B~\cite{sullivan2012overview}, which contain high-quality videos with resolution 1920$\times$1080.
	\textit{For video recognition task},  we train and evaluate it on two large scale datasets requiring temporal relation reasoning, \textit{i.e.}, {{Something V1\&V2}}~\cite{goyal2017something}\cite{mahdisoltani2018fine}.

 \begin{figure*}[!thbp]
 	\vspace{-3mm}
	\newcommand{\imgwid}{0.99}
	\centering
	\includegraphics[width=0.98 \linewidth]{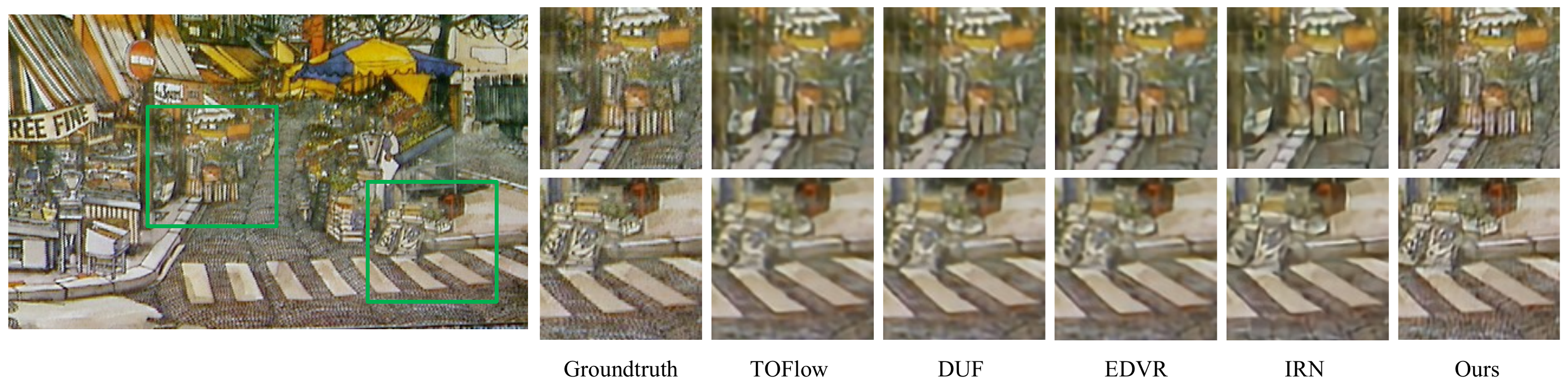}
	\vspace{-2mm}
	\caption{ Qualitative comparison on the reconstruction of 4$\times$ downscaled \textit{calendar} clip.
		\textit{Best to view by zooming in.}}
	\label{fig:qua_vid4}
	\vspace{-2mm}
\end{figure*}
	
	\begin{table*}[!thbp]
		\centering
		\scalebox{0.80}{
			\tabcolsep=0.9mm	
			\begin{tabular}{|l|l||c|c|c||c|c|c|c|c||c|}
				\hline
				Downscaling & Upscaling  &\#Frame &FLOPs &\#Param.  &Calendar-Y &City-Y &Foliage-Y  &Walk-Y &Average-Y  &Average-RGB
				\\
				\hline \hline  	
				Bicubic&Bicubic & 1 &N/A &N/A &18.83/0.4936  &23.84/0.5234	&21.52/0.4438  &23.01/0.7096  & 21.80/0.5426   &20.37/0.5106	
				\\	
				Bicubic&SPMC
				~\cite{tao2017detail}  & 3 & - & -   &~~-/- ~ &~~-/- ~ &~~-/- ~ &~~-/- ~ &25.52/0.76  &-/-
				\\
				Bicubic&Liu
				~\cite{liu2013bayesian} & 5 &- &- &21.61/-  &26.29/-  &24.99/-  &28.06/-  &25.23/-  &-/-
				\\
				Bicubic&TOFlow~\cite{xue2019video} & 7 &0.81T &1.41M &22.29/0.7273  &26.79/0.7446 &25.31/0.7118 &29.02/0.8799 &25.85/0.7659 &24.39/0.7438	
				\\
				Bicubic&FRVSR~\cite{sajjadi2018frame} & 2 &0.14T &5.05M &22.67/0.7844  &27.70/0.8063 &25.83/0.7541   &29.72/0.8971 &26.48/0.8104 &25.01/0.7917
				\\
				Bicubic&DUF-52L~\cite{jo2018deep} & 7  &0.62T &5.82M  &24.17/0.8161 &28.05/0.8235 &26.42/0.7758 &30.91/0.9165	&27.38/0.8329 &25.91/0.8166
				\\
				Bicubic&RBPN~\cite{haris2019recurrent} & 7 &9.30T &12.2M &24.02/0.8088 &27.83/0.8045 &26.21/0.7579 &30.62/0.9111 &27.17/0.8205  &25.65/0.7997 
				\\
				Bicubic&EDVR-L
				~\cite{wang2019edvr} & 7 &0.93T &20.6M  &24.05/0.8147 &28.00/0.8122	 &26.34/0.7635  &{31.02}/0.9152  &27.35/0.8264  &25.83/0.8077   	
				\\	
				Bicubic&PFNL
				~\cite{yi2019progressive} &7 &0.70T &3.00M  &23.56/0.8232 &28.11/0.8366 &26.42/0.7761 &30.55/0.9103 &27.16/0.8365 &25.67/0.8189 
				\\
				Bicubic&RLSP~\cite{fuoli2019efficient} & 3 &0.09T &4.21M  &24.36/0.8235  &28.22/0.8362 &26.66/0.7821 &30.71/0.9134 &27.48/0.8388 &25.69/0.8153
				\\
				Bicubic&TGA~\cite{TGA} &7 &0.23T &5.87M  &{ 24.50}/0.8285 &28.50/0.8442 &26.59/0.7795 &30.96/0.9171 &27.63/0.8423 &26.14/0.8258
				\\
				Bicubic&RSDN 9-128~\cite{RSDN} & 2  &0.13T &6.19M &{24.60}/{0.8355}	 & {29.20}/{0.8527}  &{26.84}/{0.7931}  &{31.04}/{0.9210} &{27.92}/{0.8505}  &{26.43}/{0.8349}
				\\
				IRN&IRN~\cite{xiao2020invertible} & 1 &0.24T &4.36M  & 26.62/0.8850  &33.48/0.9337 &29.71/0.8871 &35.36/0.9696 &31.29/0.9188 &29.21/0.8990  \\
				\hline
				SelfC-\textit{small}&SelfC-\textit{small} & 7 &{0.043T} &{1.76M}  &27.10/0.9020 & 33.49/0.9379 &30.46/0.9138 &35.40/0.9730  &31.61/0.9317 &29.64/0.9151 \\
				\textbf{SelfC-\textit{large}}&\textbf{SelfC-\textit{large}} & 7 &0.084T &3.37M &\textbf{27.82/0.9167} &\textbf{34.14/0.9454}  &\textbf{30.95/0.9210}  &\textbf{36.28/0.9778}    &\textbf{32.30/0.9402} &\textbf{30.27/0.9245}  \\
				\hline
			\end{tabular}
		}
		\vspace{-2mm}
		\small
		\caption{Quantitative comparison (PSNR(dB) and SSIM) on Vid4 for $4\times$ video rescaling.
					 Y indicates the luminance channel. 
			FLOPs (MAC) are calculated by upscaling an LR frame of size 180$\times$120.
		}
		\label{tab:sota_sr4}
			\vspace{-1mm}
	\end{table*}
	
	\begin{table*}[!thbp]
		
		\centering
		\scalebox{0.730}
		{	
			\tabcolsep=0.8mm
			\begin{tabular}{|l||c|c|c|c|c|c|c|c||c|c|}
				\hline 
				Downscaling & Bicubic   & Bicubic  &  Bicubic    &   Bicubic  &  Bicubic  & Bicubic   & Bicubic & IRN~\cite{xiao2020invertible} &SelfC-\textit{small} &\textbf{SelfC-\textit{large}} \\
				\hline  	
				Upscaling &  Bicubic   &  TOFlow~\cite{xue2019video} & FRVSR~\cite{sajjadi2018frame}  &  DUF-52L~\cite{jo2018deep}    &   RBPN~\cite{haris2019recurrent}   &  PFNL~\cite{yi2019progressive}     & RSDN 9-128~\cite{RSDN} & IRN~\cite{xiao2020invertible} &SelfC-\textit{small} &\textbf{SelfC-\textit{large}}
				\\
				\hline
				\hline
				\textit{{SPMCs-30}} &23.29/0.6385    &27.86/0.8237 &28.16/0.8421 &29.63/0.8719 &29.73/0.8663 &29.74/0.8792     & -/- &36.24/0.9559 &37.20/0.9681&\textbf{38.32/0.9744}
				\\
				\hline
				\textit{{Vimeo90K-T}} 	 &31.30/0.8687 &34.62/0.9212 &35.64/0.9319  &36.87/0.9447 &37.20/0.9458 &-/-     & 37.23/0.9471 & 40.83/0.9734 &40.68/0.9756&\textbf{41.53/0.9786}
				\\ 
				\hline 
				
			\end{tabular}
		}
		\vspace{-2mm}
		\small
		\caption{
			Quantitative comparison (PSNR-Y(dB) and SSIM-Y) on SPMCs-30 and Vimeo90K-T for $4\times$ video rescaling.
		} % Best view in color.
		\label{tab:sota_other_ds}
		\vspace{-5mm}
	\end{table*}

	\vspace{\subsecspace}
	\subsection{Implementation Details}
	\vspace{\subsecspace}
	\textbf{Video rescaling.} $\lambda_1$, $\lambda_2$, $\lambda_3$ and $\lambda_4$ are set as 0.01, 1, 1 and 1, respectively. $\lambda_1$ is decayed to 0 in the last 200K iterations.
	Each training clip consists of 7 consecutive RGB patches of size 224$\times$224.
	The batch size is set as 8. We augment the training data with random horizontal flips and 90$^{\circ}$ rotations. We train our model with Adam optimizer~\cite{kingma2014adam} by setting $\beta_1$ as 0.9, $\beta_2$ as 0.999, and learning rate as $10^{-4}$. The total training iteration number is about 400K. The learning rate is divided by 2 every 100K iterations. We implement the models with the PyTorch framework~\cite{paszke2019pytorch} and train them with 2 NVIDIA 2080Ti GPUs.
	We draw 5 times from the generated distribution for each evaluation and report the averaged performance. We leverage the invertible neural network (INN) architecture to implement the CNN parts of the paired frequency analyzer and synthesizer for a fair comparison with IRN because INN will cut the number of parameters by 50\%.
	We propose the two following models: the SelfC-\textit{small} and SelfC-\textit{large}, which consist of 2 and 8 invertible Dense2D-T blocks. The invertible architecture follows~\cite{xiao2020invertible}.
	Training the SelfC-\textit{large} model takes  $\sim$6days.

	\textbf{Video compression.} The rescaling ratio of the SelfC is set to 2. We adopt H.265 as the embedded codec. $\lambda_4$ is set as 0.1 to make sure the statistical distribution of the downscaled videos is more closed to the natural images, which stabilizes the performance of the whole system.
	$\lambda_{codec}$ is empirically set as 4. 
 	The CRF value of the H.265 codec is randomly selected from \{11,13,17,21\} during training while it is set to a fixed value during evaluation. The input video clip length is set as 3.
	The other details follow that of video rescaling task. The models are initialized from SelfC-\textit{large} model but the number of invertible Dense2D-T blocks is reduced to 4. The surrogate CNN is randomly initialized, and is jointly optimized with other components of SelfC online. It takes about 5days to train the model.
	
	\textbf{Action recognition.}~We insert the downscaler of our framework before the action recognition CNN, \textit{i.e.}, TSM~\cite{lin2019tsm}.
	 The data augmentation pipeline also follows it.
	 The downscaling ratio of SelfC is 2. At inference time, we used just {1} clip per video and each clip contains {8} frames. We adopt 2 plain Dense2D-T blocks as the CNN part of the frequency analyzer. Note that the downscaler is first pretrained on Vimeo90K dataset with the rescaling task.

	\vspace{\subsecspace}
	\subsection{Results of Video Rescaling}
	\vspace{\subsecspace}
	As shown in Tab.~\ref{tab:sota_sr4} and Tab.~\ref{tab:sota_other_ds},
	our method outperforms the recent state-of-the-art video super resolution methods on Vid4, SPMCs-30 and Vimeo90K-T by a large margin in terms of both PSNR and SSIM. For example, the result on the Vid4 dataset for our SelfC-large model is 32.85dB, while the state-of-the-art video super resolution approach RSDN only achieves 27.92dB.
	Furthermore, we also provide the results of the image rescaling method, \textit{i.e.}, IRN, in Tab.~\ref{tab:sota_sr4} and Tab.~\ref{tab:sota_other_ds}. 
	It is obvious that our method outperforms IRN significantly while also reduces the computational complexity by $\sim$3$\times$ (SelfC-\textit{large}) or 6$\times$ (SelfC-\textit{small}). 
	This result clearly proves that it is necessary to exploit the temporal relationship for video rescaling task, while the existing image rescaling methods like IRN ignored this temporal cue. 
	We also show the qualitative comparison with other methods on Vid4 in Fig.~\ref{fig:qua_vid4}.
	Our method demonstrates much better details than both video super-resolution methods and image rescaling methods, proving the superiority of the video rescaling paradigm.

	\vspace{\subsecspace}
	\subsection{Results of Video Compression}
	\vspace{\subsecspace}
	In our framework, we adopt the H.265 codec in \textit{default}\footnote{\textit{ffmpeg -pix\_fmt yuv444p -s WxH -r 50 -i video.yuv -c:v libx265 -preset veryfast -x265-params ``qp=Q''}} or \textit{zerolatency}\footnote{\textit{ffmpeg -pix\_fmt yuv444p -s WxH -r 50 -i video.yuv -c:v libx265 -preset veryfast -tune zerolatency -x265-params ``qp=Q''}} modes.
	The first mode is for the offline storage of video data while the other one is oriented to the online low-latency video streaming services.
	The evaluation metrics are PSNR and MS-SSIM~\cite{wang2003multiscale}.

\begin{table}[!b]
		\vspace{-4mm}
	\centering
	\small
	\scalebox{0.92}
	{
		\tabcolsep=2.1mm
		\begin{tabular}{|c|c|c|c|c|}
			\hline
		     \multirow{2}{*}{H.265 mode}	& \multicolumn{3}{c|}{BDBR-SSIM}& \multicolumn{1}{c|}{BDBR-PSNR} \\
			\cline{2-5}
			 & UVG & MCL-JCV& HEVC B &  UVG \\ 
			\hline
			{\tabincell{c}{\textit{default} }}  &-50.84  &-36.48 & -30.61 & -49.01 \\
			\hline
			{\tabincell{c}{\textit{zerolatency} }}&-33.98  &-28.16 & -18.04 & -19.41  \\
			\hline
		\end{tabular}
	}
	\vspace{-1mm}
	\caption{
		BDBR results using H.265 as the anchor.
		The lower value, the more bit cost reduced.
	}
	\label{tab:bdbr}
	\vspace{-3mm}
\end{table}

			\begin{figure*}[!thp] 
		\centering
		\newcommand{\widthscalefive}{0.25}
		\tabcolsep = 0.1mm
		\scalebox{0.97}{
			\begin{tabular}{cccc}
				\includegraphics[width=0.24 \textwidth]{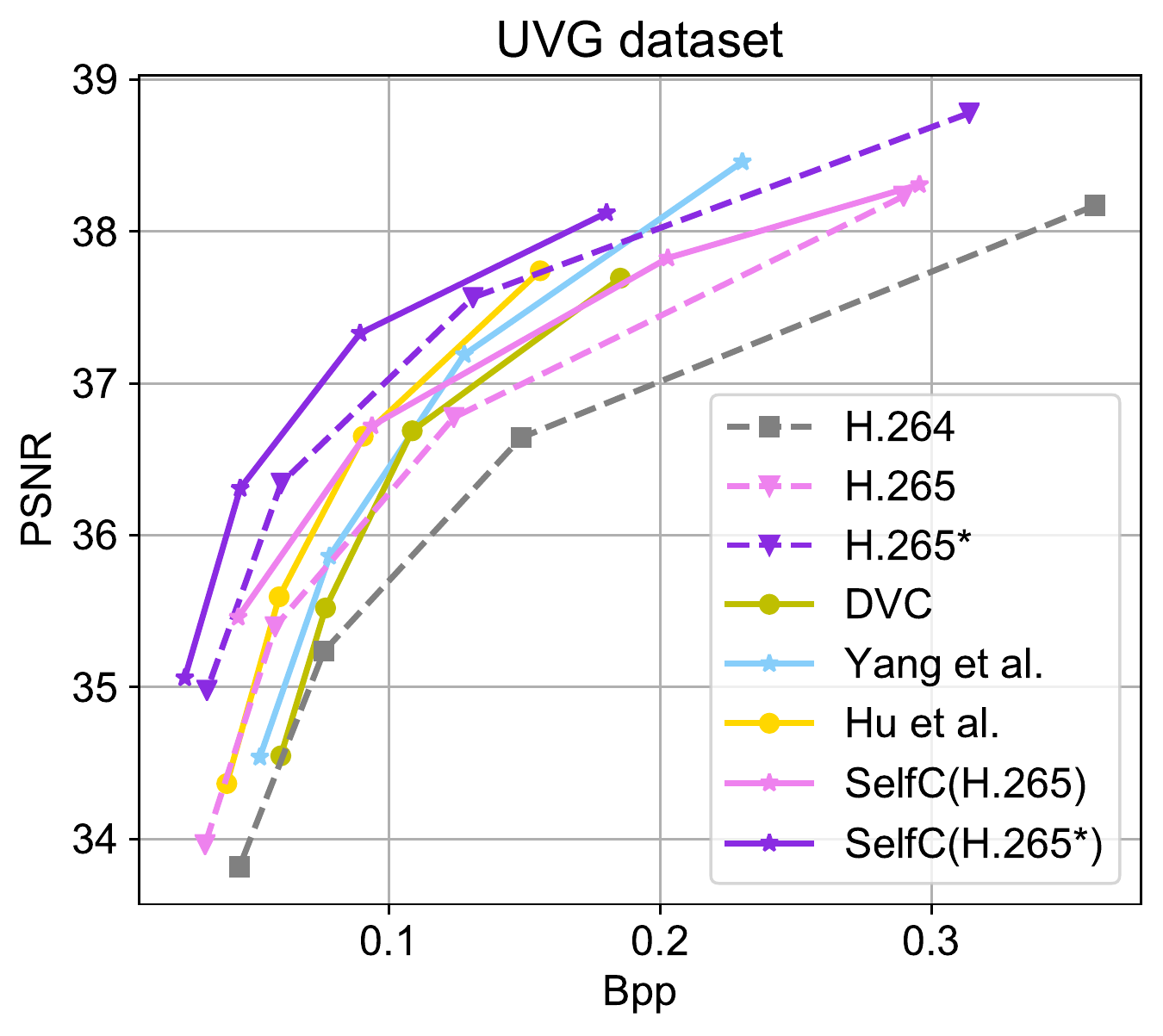} & 
				\includegraphics[width=0.26 \textwidth]{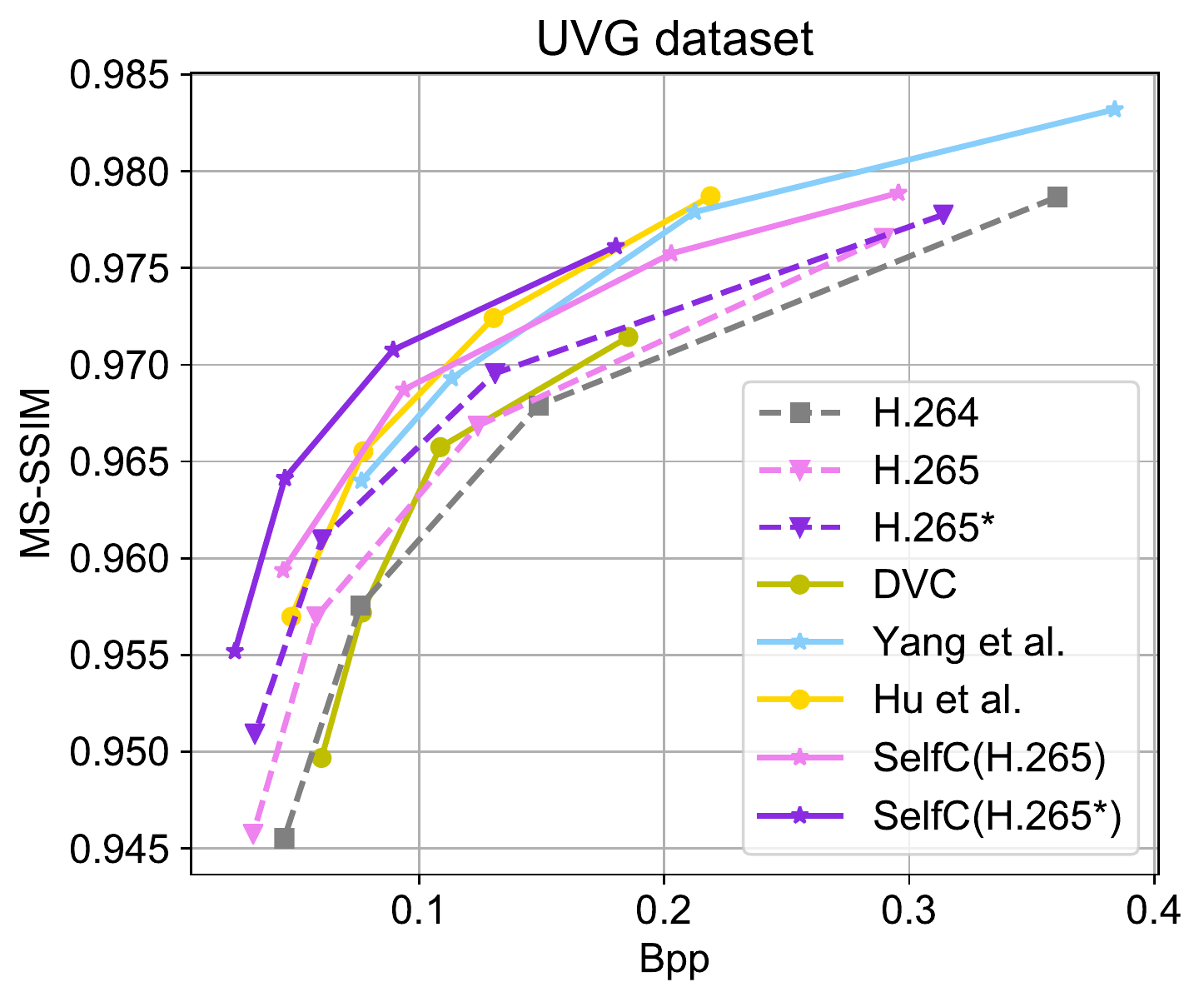} &
				\includegraphics[width=0.26 \textwidth]{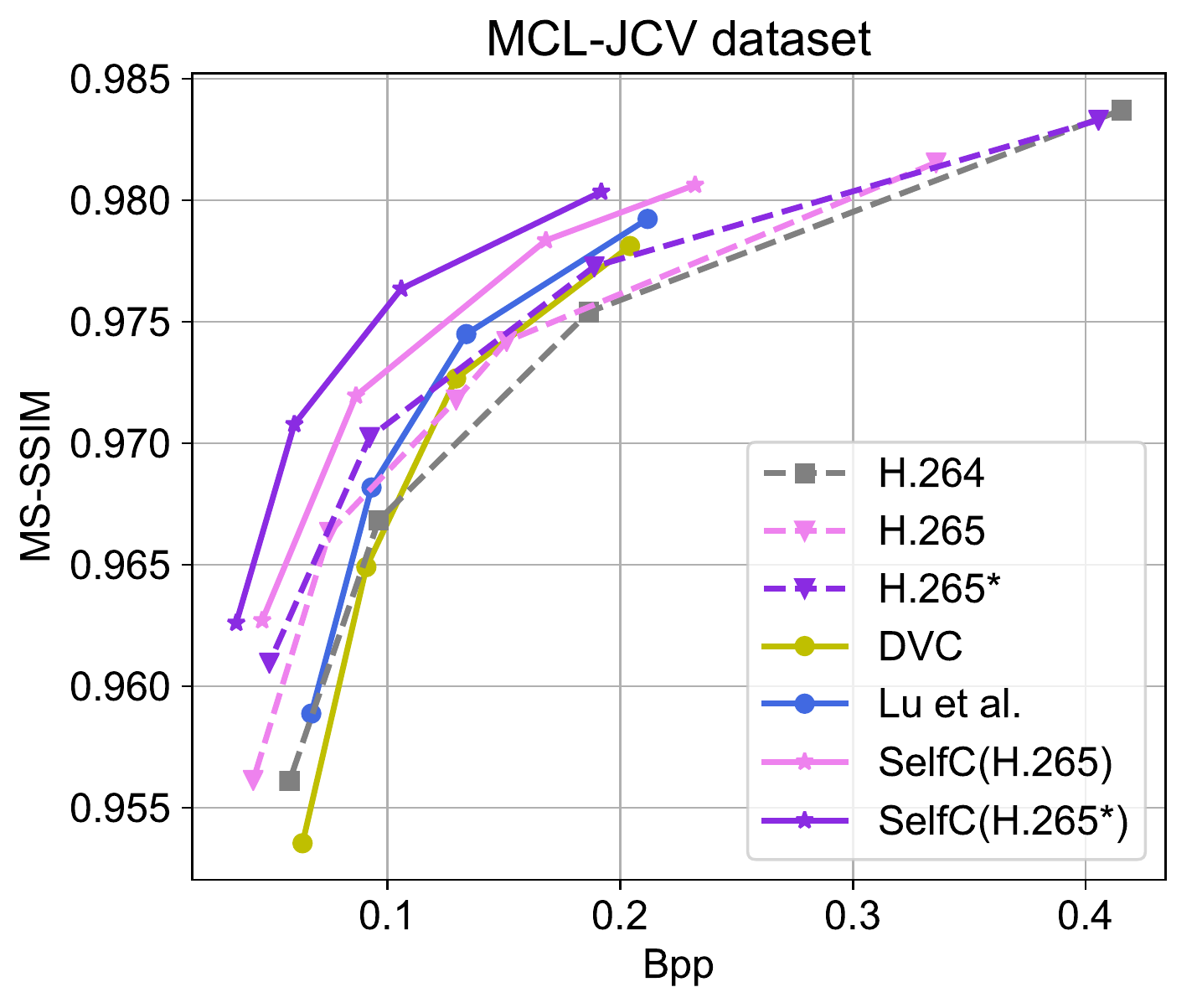} & 
				\includegraphics[width=0.253 \textwidth]{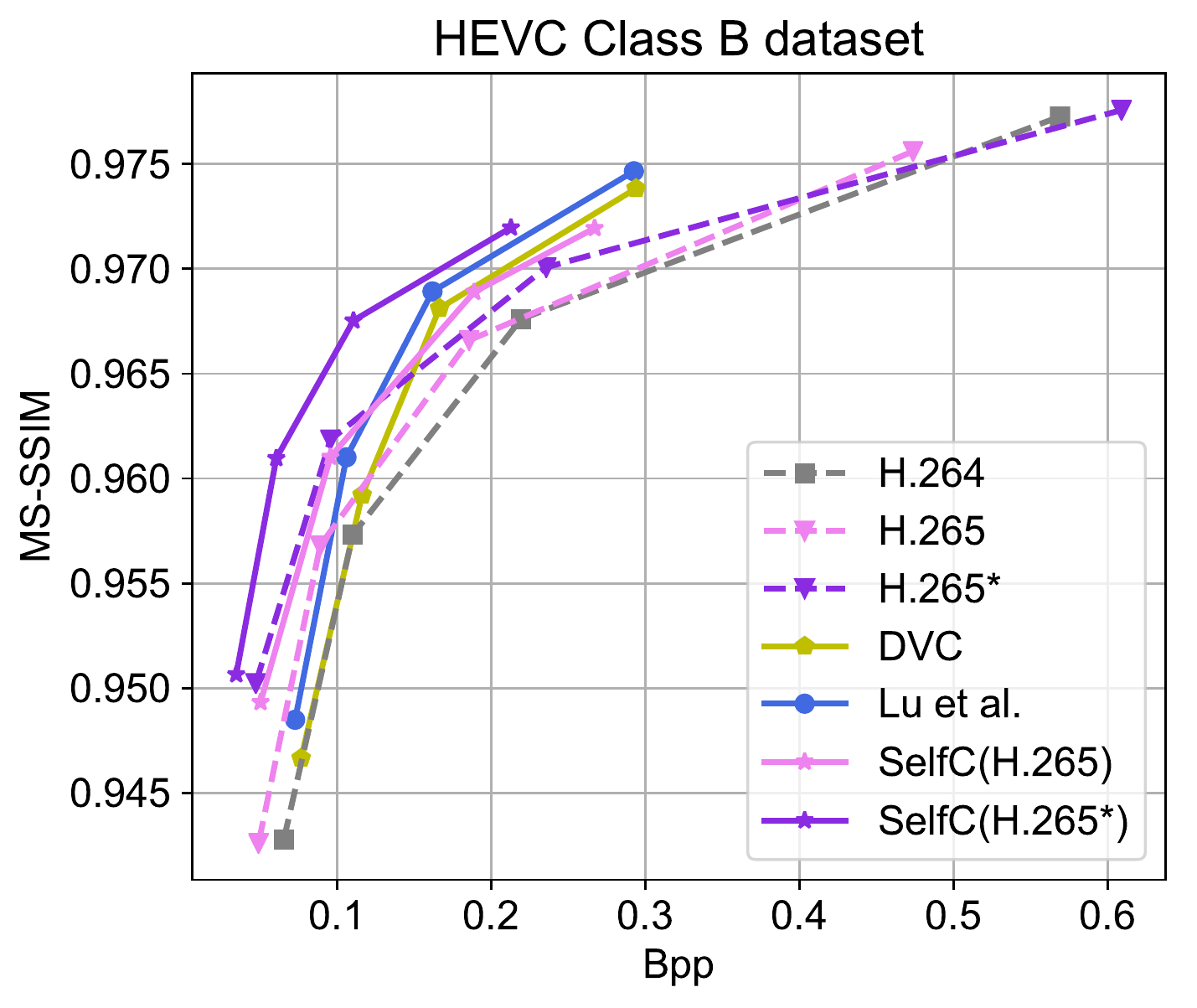} 
			\end{tabular}
		}
		\vspace{-3mm}
		\caption{
			Comparison between the proposed method with H.265, H.264, and learnable video codecs. H.265 and H.265* denote the H.265 codec in \textit{zerolatency} and \textit{default} modes, respectively.
		}
		\vspace{-3mm}
		\label{fig:compression_sota}
	\end{figure*}

	Fig.~\ref{fig:compression_sota} shows the experimental results.
	It is obvious that our method outperforms both the traditional methods and learning-based methods (DVC~\cite{lu2019dvc}, Yang~\etal~\cite{yang2020learning}, Hu~\etal~\cite{hu2020improving}, and Lu~\etal~\cite{lu2020content}) on video compression task by a large margin.
	Although our method is only optimized by $\ell_1$ loss, it demonstrates strong performances in terms of both PSNR and MS-SSIM metrics. It should also be mentioned that our models generalize well to \textit{default} mode although only trained with the \textit{zerolatency} mode of H.265 codec.
	We also evaluate the Bjøntegaard Delta Bit-Rate (BDBR)~\cite{bjontegaard2001calculation} by using H.265 as the anchor method.
	 As shown in Tab.~\ref{tab:bdbr}, our method saves the bit cost or the storage space by over 30\% averagely under the same MS-SSIM, compared with the vanilla H.265 codec. Notably, we reduce the bit cost by over 45\% on the UVG dataset. This proves that the video rescaling technique is a novel and effective way to improve video compression performance, without considering much about the complicated details of the industrial lossy codecs.

\begin{figure}[!tp]
	%		\vspace{-4mm}
	\begin{minipage}[c]{0.23 \textwidth}
		\begin{center}
			\includegraphics[width=\textwidth]{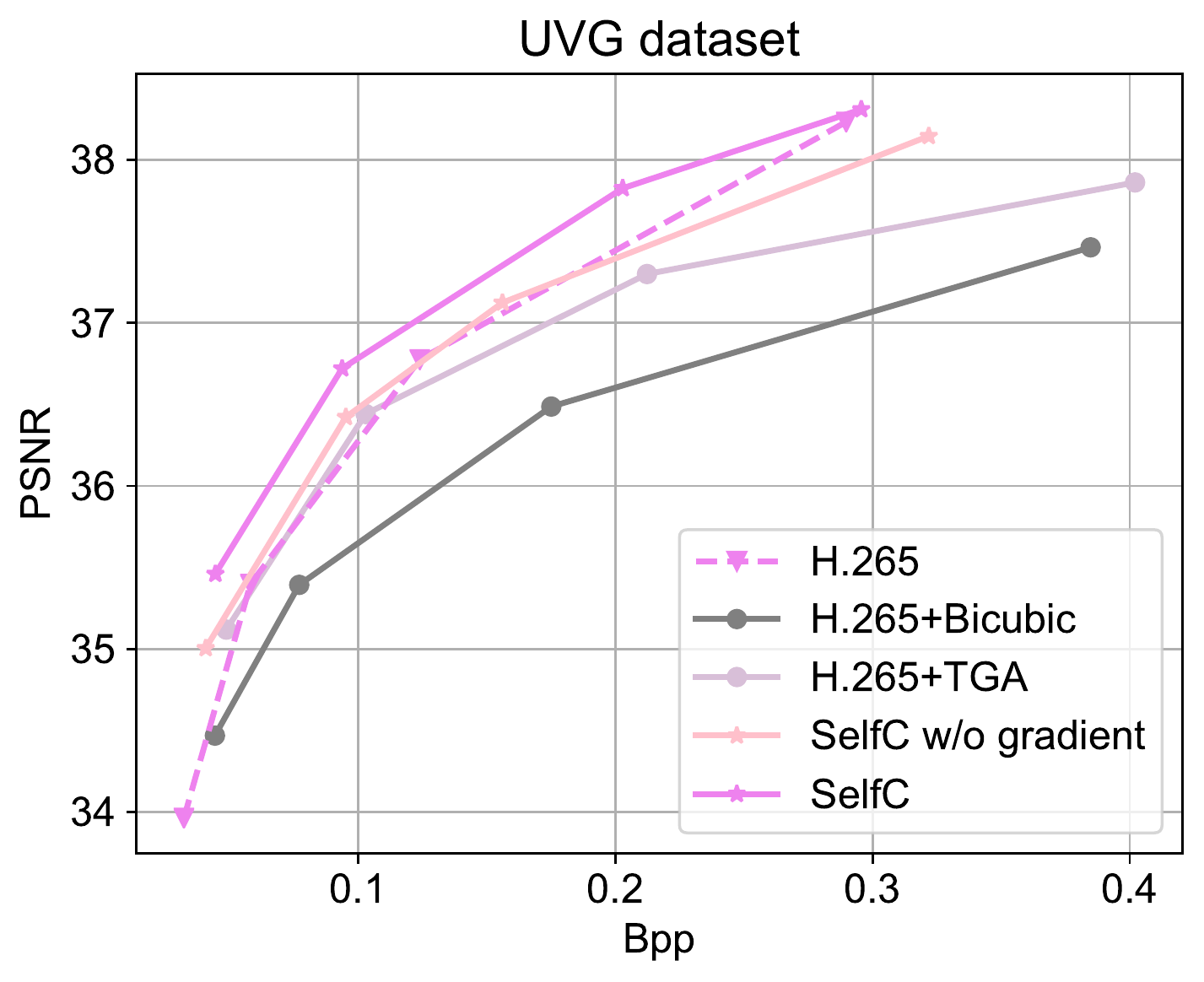}
		\end{center}
	\end{minipage}\hfill
	\begin{minipage}[c]{0.24 \textwidth}
		\caption{Comparison of our ``Video  rescaling+Codec'' scheme with other paradigms.
			SelfC w/o gradient model is trained by neglecting the H.265 codec gradients~\cite{bengio2013estimating}. 
		}
		\label{fig_compression_ab}
	\end{minipage}
	\vspace{-4mm}
\end{figure}

	We perform more analysis to verify the effectiveness of the ``Video rescaling+Codec'' paradigm and the proposed gradient estimation method. As shown in Fig.~\ref{fig_compression_ab}, it is observed that using Bicubic as the downscaler and upscaler in the video compression system (\textit{i.e.,} H.265+Bicubic) leads to a much inferior result than the baseline.
	We also try to improve the result by using a state-of-the-art video super resolution method, \textit{i.e.}, TGA~\cite{TGA}. The performance is indeed improved though still lower than the baseline method H.265. 
	Considering the network parameters of TGA are 5.87M while ours are only 0.88M, this result further demonstrates the effectiveness of our SelfC framework.  Finally, we provide experimental results (\textit{i.e.,} SelfC w/o gradient) when directly using the biased Straight-Through Estimator~\cite{bengio2013estimating} for approximating the gradient of H.265. The results show that the proposed gradient estimation method in Section~\ref{app1_compression} can bring nearly 0.3dB improvements.

	Finally, We give the complexity analysis of the proposed video compression system. Although it seems like that our method adds extra computation costs upon the H.265 codec, our system is indeed more efficient because the input videos to codec are downscaled. Concretely, under the \textit{zerolantency} mode, for one frame with resolution 1920$\times$1080, the average encoding time of our method is 108ms, including 21ms for 2$\times$downscaling and 87ms consumed by the embedded H.265 codec. Our system improves the efficiency of the vanilla H.265 codec (116ms) and is also $\sim$5$\times$ faster than the learnable codec DVC (522ms).
		\begin{figure*}[!th] 
		\vspace{-3mm}
		\centering
		\newcommand{\widthscalefive}{0.16}
		\renewcommand{\arraystretch}{1}
	\tabcolsep = 0.2mm
	\begin{tabular}{c@{\hskip  0.2mm}c@{\hskip  0.2mm}c@{\hskip  0.4mm}c@{\hskip 0.2mm}c@{\hskip 0.2mm}c}
		Input HR video $x$ &
		LR video $x_l$&
		$x_l - \textsf{\small Bicubic}(x)$ &
		HF component $f_h$ &
		Sample 1 of $\hat{f}_h$ &
		Sample 2 of $\hat{f}_h$
		\\
		\vspace{3mm}
		\includegraphics[width=0.159 \textwidth]{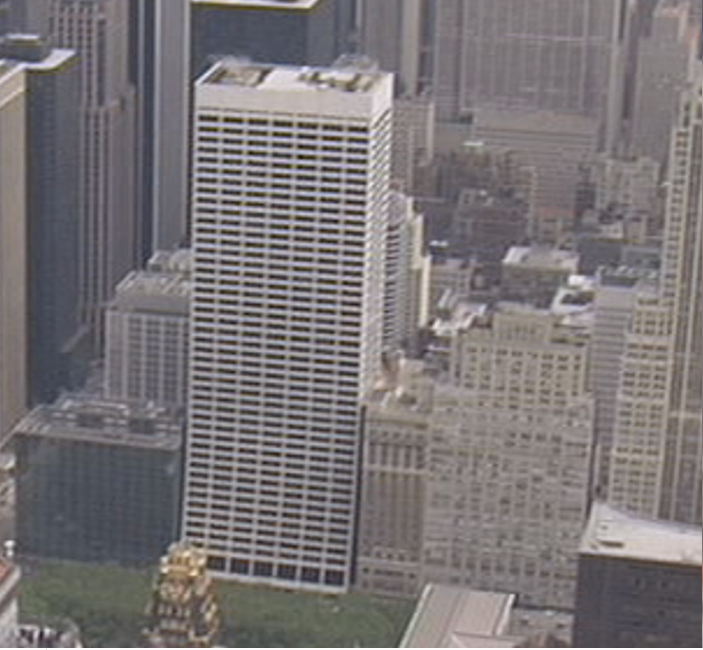} & 
		\includegraphics[width=0.162 \textwidth]{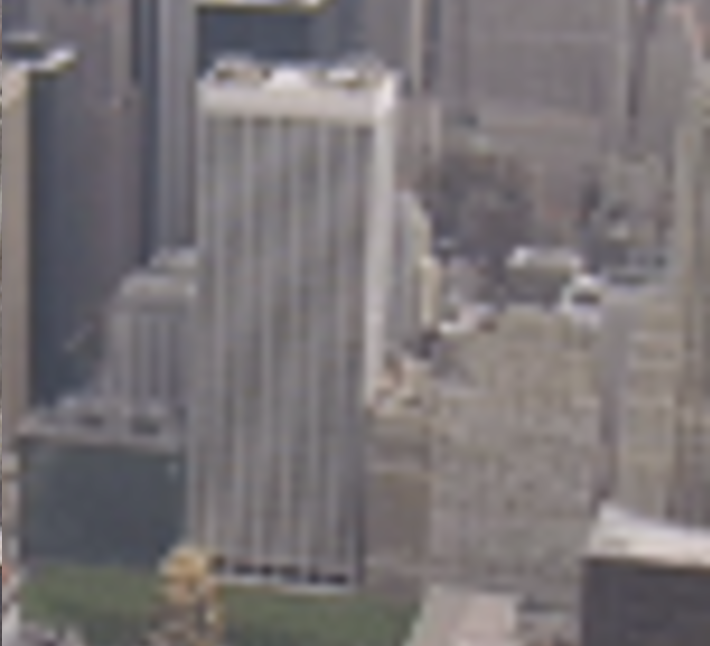} & 
		\includegraphics[width=0.159 \textwidth]{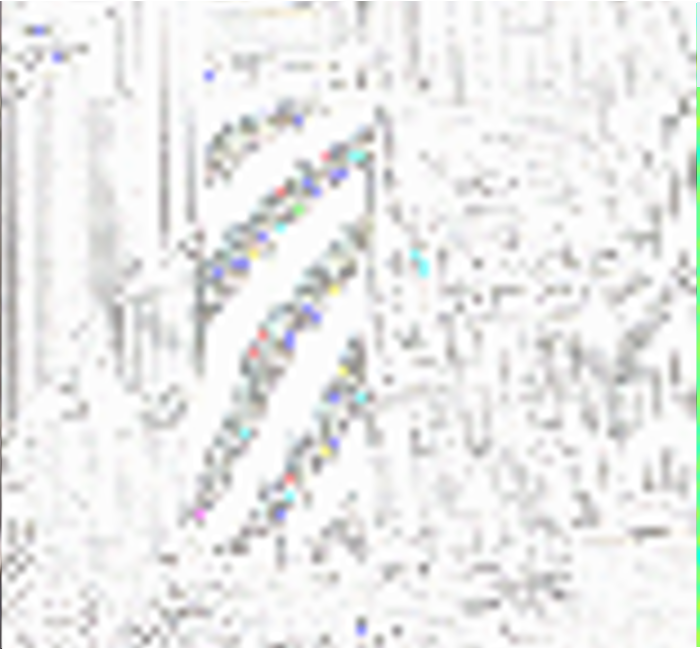} & 
		\includegraphics[width=\widthscalefive \textwidth]{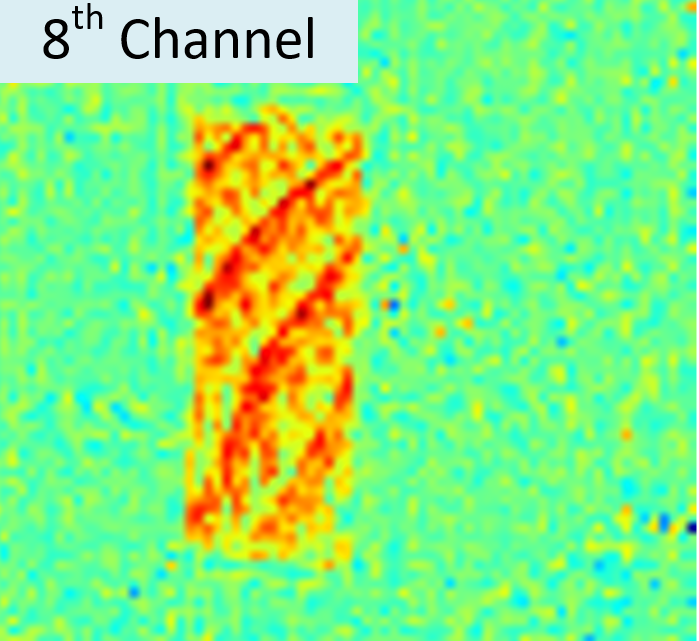} &
		\includegraphics[width=\widthscalefive \textwidth]{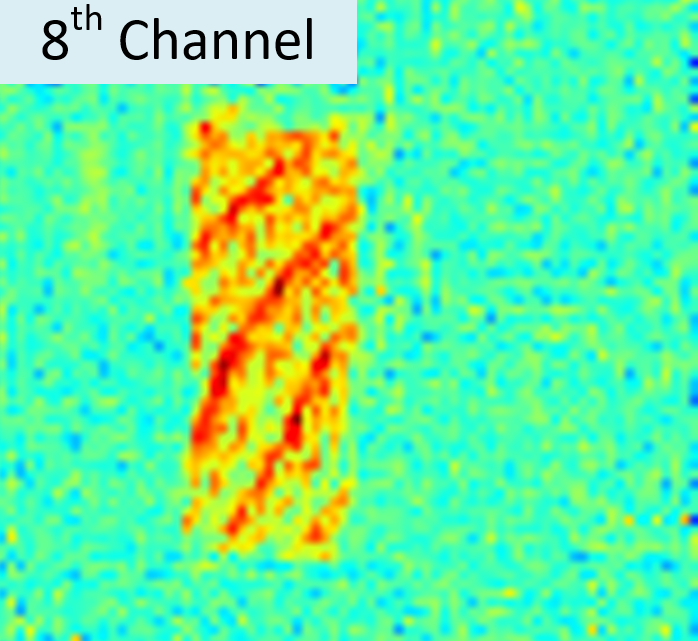} & 
		\includegraphics[width=\widthscalefive \textwidth]{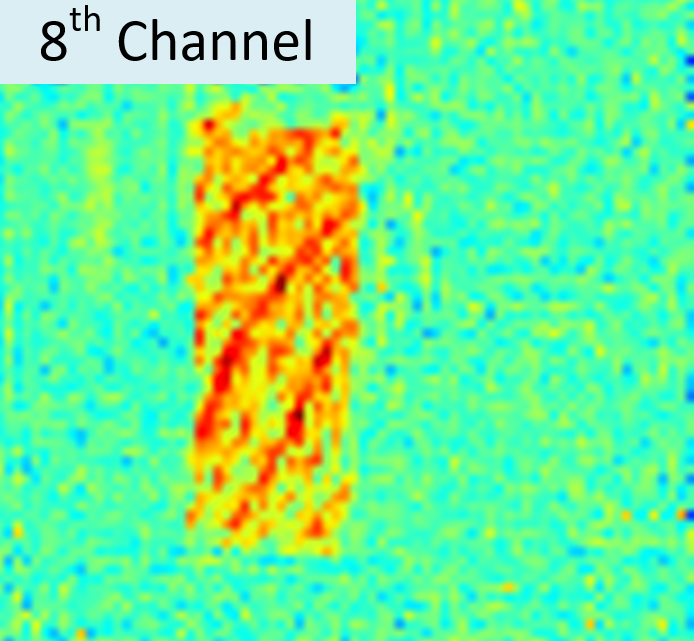} 
	\end{tabular}
		\vspace{-3mm}
		\caption{
		Visualization of the high-frequency component $\hat{f}_h$ sampled from the learned distribution $p(f_h|x_l)$.
		We also compare the difference of the downscaled video by our learnable downscaler and the Bicubic downsampler. Since the magnitude of the difference is small, we amplify it by 10 times for better visualization.
		}
		\vspace{-4mm}
		\label{fig:vis_distribution}
	\end{figure*}

	\vspace{\subsecspace}
	\subsection{Results of Efficient Video Action Recognition}
	\vspace{\subsecspace}
	\label{sec_exp_action}

	We show the video action recognition results in Tab.~\ref{tab:ft_strategy}. In the first group, when directly testing the action model pretrained on full resolution videos, we observe the performances of low-resolution videos downscaled by Bicubic and ours are both dropped drastically, because the classification networks are rather sensitive to the absolute scale of the input. However, our downscaler still performs better (30.4\% vs. 32.7\% on Something V1 dataset).

	In the second group, we provide the experimental results when the action recognition CNN is fine-tuned on the low-resolution videos downsampled by Bicubic and our downscaler.
	It is obvious that our method clearly outperforms Bicubic downsampler by about $1.5\%$ in terms of Top1 accuracy on the two datasets. Notably, our downscaler is learnable. Therefore, we then jointly fine-tune action recognition CNNs and our downscaler. The results in the last group show that the end-to-end joint training strategy further improves the performance by an obvious margin. On Something V2, the ultimate performances of our method nearly achieve that of performing recognition directly on HR videos, and our method improves the efficiency by over 3 times.
	The downscaler of IRN can not improve the efficiency of this task because its computation cost is even larger than the HR setting. We try to decrease the layer number of IRN but it no longer converges.

		\begin{table}[!hbp]
%	\vspace{0mm}
%\vspace{-2mm}
		\centering
		\tabcolsep=0.9mm
		\renewcommand{\arraystretch}{0.95}
		\scalebox{0.79}{
			\begin{tabular}{|c|c|c|c|c|c|c|c|}
				
				\hline
				\multirow{2}{*}{\tabincell{c}{Method} } & 
				\multirow{2}{*}{\tabincell{c}{Action \\CNN} } & 
				\multirow{2}{*}{\tabincell{c}{FLOPs} }&
				\multirow{2}{*}{\tabincell{c}{Params} } &
				\multicolumn{2}{c|}{V1} &  
				\multicolumn{2}{c|}{V2} \\
				\cline{5-8}	&&&& \scriptsize{Top1 (\%)} & \scriptsize{Top5 (\%)} &\scriptsize{Top1 (\%)} & \scriptsize{Top5 (\%)}  \\
				
				\hline\hline
				HR &TSM & 33.0 G    & 24.31M      & 45.6 & 74.2 &58.8 & 85.4 \\
				\hline
				Bicubic          &TSM & 8.6G & 24.31M    		   & 30.4 & 57.5 & 40.1 & 68.7 \\
				Ours            &TSM & 10.8G & 24.35M    		   & 32.7 & 59.5 & 41.8 & 71.5 \\
				\hline
				Bicubic (\textsf{\small FT})          &TSM & 8.6G & 24.31M    		   & 42.1 & 72.1 & 56.0 & 83.9 \\
				Ours (\textsf{\small FT})            &TSM   & 10.8G & 24.35M 		   & 43.5 & 72.8   &57.4 & 84.8 \\
				\hline
				Ours (\textsf{\small E2E}) & TSM        & {10.8G} & 24.35M& \textbf{44.6} & \textbf{73.3}   & \textbf{58.3} & \textbf{85.5} \\
				\hline
				
			\end{tabular}
		}
		\vspace{-2mm}
		\caption{
			Comparison between our method and Bicubic downscaler.
			\textsf{\small FT} denotes only fine-tuning the action recognition CNN.
			\textsf{\small E2E} denotes also fine-tuning the dowscaler.
		}
		\label{tab:ft_strategy}
		\vspace{-3mm}
	\end{table}

		\subsection{Ablation Studies on the Framework}
	In this section, we conduct experiments on video rescaling task to verify the effectiveness of the components in our framework. We first define the following 2 baselines:
	(1) {IRN}~\cite{xiao2020invertible}, which is the most recent state-of-the-art image rescaling method. For fair comparison, we re-train it on Vimeo90K dataset using the codes open-sourced by the authors.
	(2) {{Auto-Enc}}, which is a simple auto encoder-decoder architecture by removing the STP-Net of our model.
%	, where the low-resolution video is the latent representation.
%	We also designs several variants degraded from our model to verify the importance of temporal modeling and conditioned distribution modeling.
	The experimental results are shown in Tab.~\ref{tab:abl_exp}.
	
	\renewcommand{\arraystretch}{0.95}
\begin{table}[!thbp]
	%	\begin{minipage}{0.6\linewidth}
	\centering
	\small
	\scalebox{0.86}
	{
		\tabcolsep=0.5mm
		\begin{tabular}{|c|c|Hc|HHc|c|c|c|}
			\hline
			\multirow{2}{*}{\tabincell{c}{Methods} }           &
			\multirow{2}{*}{\tabincell{c}{Backbone} } &
			\multirow{2}{*}{\tabincell{c}{Temporal                                                                                                                                                         \\feature?} }&
			\multirow{2}{*}{\tabincell{c}{Probability \\model} } &
			\multirow{2}{*}{\tabincell{c}{ T-prior?} } &
			\multirow{2}{*}{\tabincell{c}{Self-conditioned?} } &
			
			\multirow{2}{*}{\tabincell{c}{Param(M)} }     &
			\multicolumn{2}{c|}{Vid4-Y}                                                                                                                                                              \\
			\cline{8-9}                                        &                  &        &                                &        &        &               & \scriptsize{PSNR(dB)} & \scriptsize{SSIM} \\
			\hline\hline
	
			Auto-Enc                                            & 16$\times$Dense2D-T    & \xmark & - & \xmark & \xmark & 3.63          & 28.91       & 0.8797   \\
	
			IRN$^*$                                           & 16$\times$Dense2D    & \xmark & \tabincell{c}{Normal} & \xmark & \xmark & 4.36          & 30.68         & 0.9067   \\
			IRN-T                                            & 16$\times$Dense2D-T    & \xmark & \tabincell{c}{Normal} & \xmark & \xmark & 3.63          & 30.42        & 0.9004   \\
		
			\hline
			SelfC-\textit{basic}                                           & 2$\times$Dense2D     & \xmark & GMM(K=1)                       & \cmark & \cmark & 1.77          & 30.62                  & 0.9214            \\
			SelfC-\textit{basicT}                                              & 2$\times$Dense2D-T   & \cmark & GMM(K=1)                       & \cmark & \cmark & 1.61          & 31.29                  & 0.9268            \\
			{SelfC-\textit{small}}                      & 2$\times$Dense2D-T   & \cmark & GMM(K=5)                       & \cmark & \cmark & 1.76 & 31.61         & 0.9317   \\
			\textbf{SelfC-\textit{large}}                      & 8$\times$Dense2D-T   & \cmark & GMM(K=5)                       & \cmark & \cmark & 3.37          & \textbf{32.30}         & \textbf{0.9402}   \\

			\hline
		\end{tabular}
	}
	\vspace{-2mm}
	\caption{
		Ablation studies on 4$\times$ video rescaling.
		All blocks adopt the INN architecture for a fair comparison with IRN.
		$^*$ indicates the model is re-trained on Vimeo90K.
	}
	\label{tab:abl_exp}
	\vspace{-2mm}
\end{table}
	%	First, the {conditional probabilistic model} and the {temporal enhanced frequency analyzer} in our method both improve the performance.
	First, the Auto-Enc baseline shows more inferior performance than both IRN and our method. This proves that explicitly modeling the lost information is important. IRN is inferior to SelfC-\textit{small} model although IRN leverages an {8} times heavier backbone.
	We also tried to equip IRN with the temporal modeling ability by replacing its backbone from Dense2D to Dense2D-T. Surprisingly, the performance of the resulted model IRN-T decreases by 0.26dB. The reason is that IRN relies on the complex non-linear transformation to transform the real distribution of the lost information to the normal distribution while the transformation ability of the Dense2D-T is weaker (missing {0.73M} parameters).

	For our method, we start from the most simple model denoted by SelfC-\textit{basic}, where the backbone consists of only spatial convolutions, and the STP-Net only outputs a simple Gaussian distribution. The performance of this model is comparable with IRN but with 2$\times$ fewer parameters. This proves the efficiency and superiority of the proposed self-conditioned distribution modeling scheme.
	Then, we introduce an improved model denoted by SelfC-\textit{basicT}. The temporal modeling ability of the model is enhanced by changing the basic block from Dense2D to Dense2D-T. This leads to {0.67dB} improvement while reducing the parameters, proving the effectiveness of the Dense2D-T block for video tasks.
	Further, we increase the mixture number of the GMM model to 5. The resulted SelfC-\textit{small} model outperforms all the baselines by a large margin with only {1.76M} parameters. Our model is also scalable with larger backbone network. Enlarging the backbone by 4 times further improves the performance by 0.69dB. 
%	For more ablation studies on the depth of backbone network, comparison of different probabilistic modeling methods, the architecture of the STP-Net and the loss functions, please refer to the supplementary material.

	\vspace{\subsecspace}
	\subsection{Visualization Results}
	\vspace{\subsecspace}
	While the previous quantitative results validate the superiority of the proposed self-conditioned modeling scheme on several tasks, it is interesting to investigate the intermediate components output by our model, especially the distribution of the high-frequency (HF) component predicted by STP-Net.
	Note that the distribution is a mixture of Gaussian and includes multiple channels,
	we draw two samples of $\hat{f}_h$ from $p(f_h|x_l)$ and randomly select 1 channel of them for visualization. The $f_h$ output from the frequency analyzer is adopted as the ground-truth sample.
	
	As shown in Fig.~\ref{fig:vis_distribution}, we first see that the LR video $x_l$ downscaled by our method is modulated into some mandatory information for reconstructing the HF components more easily, compared to Bicubic. Also, the sampled HF components can restore the ground-truth of that accurately in terms of key structures, \textit{i.e.}, the windows of the building, while retaining a certain degree of randomness. This is consistent with our learning objectives.

	%%%%%%%%%%%%%%%
	\section{Conclusion}
	We have proposed a video-rescaling framework to learn a pair of downscaling and upscaling operations.
	Extensive experiments demonstrated that our method can outperform the previous methods with a large margin while with much fewer parameters and computational costs.
	Moreover, the learned operators facilitates the tasks of video compression and efficient action recognition significantly.
	
	\noindent \textbf{Acknowledgement} This work was supported by the National Science Foundation of China (61831015, 61527804 and U1908210).
	
	{\small
		\bibliographystyle{ieee_fullname}
		\bibliography{egbib}
	}
	\newpage
	\section{Appendix}
	
	Compared with the initial submission, we get better performance for both re-trained IRN~\cite{xiao2020invertible} model and our rescaling models in this submission.
	The reason is that we adopt the following improved training strategies:
	(1) When generating the LR images, we adopt the Gaussian blur downsampling strategy~\cite{RSDN} instead of the Bicubic downsampling operator provided by the Pytorch framework. Specifically, in this submission, the corresponding low-resolution patches are obtained by applying Gaussian blur with $\sigma$ = 1.6 to the HR patches followed by 4$\times$ downsampling.
	(2) In this submission, we also adopt larger HR patches of size 224$\times$224 in the first training stage. While in the initial submission, to reduce the training time, we use the HR patches of size 72$\times$72 at the first 150,000 training iterations. 
	(3) We increase the training iteration number from 240,000 to 400,000. And the decay strategy of the learning rate is also smoother.
	
	For the video compression part, we reduce the input clip length of the SelfC framework to 3 in this submission. With this smaller clip length, the training procedure can be conducted with only 2 Nvidia 2080Ti GPUs, which makes our work easier to follow. We give more details about the settings for the evaluation in the main paper.
	
	We also try to tune other hyper-parameters for a more stable training process.
	
		\subsection{More Details about the Basic Blocks}
\begin{figure}[!thbp]
		\begin{center}
			\includegraphics[width=0.75\linewidth]{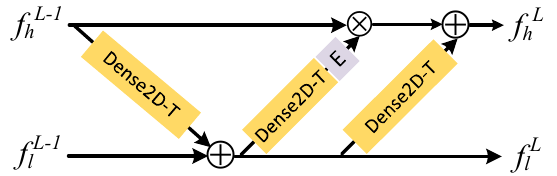}
			\caption{
				Invertiable Dense2D-T block.
				$\oplus$ and $\otimes$ denote the element-wise addition and element-wise multiplication respectively.
				``E'' denotes the exponential function $exp(\cdot)$.
			}
			\vspace{-10mm}
			\label{fig:inv_dense2d_t}
		\end{center}
	\end{figure}

\begin{figure}[!thbp]
		\begin{center}
			\includegraphics[width=0.75\linewidth]{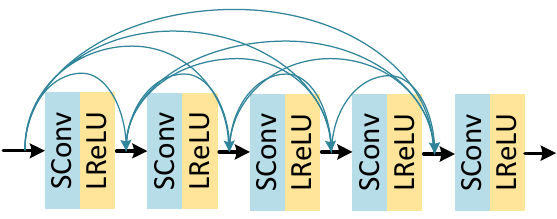}
			\caption{
				Dense2D block. SConv denotes the spatial convolution, i.e., 3D convolution of kernel size 1$\times$3$\times$3. LReLU denotes the Leaky ReLU non-linearity.
			}
			\vspace{-10mm}
			\label{fig:dense2d}
		\end{center}
	\end{figure}

\begin{figure}[!thbp]
		\begin{center}
			\includegraphics[width=0.75\linewidth]{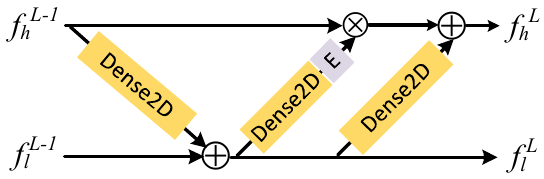}
			\caption{
				Invertiable Dense2D block.
				$\oplus$ and $\otimes$ denote the element-wise addition and element-wise multiplication respectively.
				``E'' denotes the exponential function $exp(\cdot)$.
			}
			\vspace{-10mm}
			\label{fig:inv_dense2d}
		\end{center}
	\end{figure}

	Besides the Dense2D-T blocks illustrated in Fig. 1 of the main paper, we also demonstrate the network structure of Dense2D in Fig.~\ref{fig:dense2d}. The stride size of the convolutions within the blocks above is 1.
	%	The output channel number of SConvs is 24.
	The schematic diagram for the invertiable blocks based on Dense2D and Dense2D-T blocks are shown in Fig.~\ref{fig:inv_dense2d} and Fig.~\ref{fig:inv_dense2d_t}.
	
\end{document}